\documentclass{article}

\usepackage[dblblindworkshop,final]{neurips_2026}
\workshoptitle{From Pretrained Representations to Acting Agents}

\usepackage[utf8]{inputenc}
\usepackage[T1]{fontenc}
\usepackage{url}
\usepackage{booktabs}
\usepackage{amsfonts}
\usepackage{amsmath}
\usepackage{nicefrac}
\usepackage{microtype}
\usepackage{xcolor}
\usepackage{graphicx}
\graphicspath{{figures/}}
\usepackage{float}
\usepackage{hyperref}
\hypersetup{colorlinks=true, allcolors=black}
\title{Can We Tell What GRPO Post-Training Added?\\
A Cross-Scorer Study Across Math and Code}

\author{%
  Masoud Fetanat \\
  UNSW Sydney \\
  \texttt{m.fetanat@unsw.edu.au} \\
}

\begin{document}

\maketitle

\begin{abstract}
Does the scorer change how much improvement we measure after Group Relative
Policy Optimization (GRPO)? We train \texttt{Qwen-2.5-7B-Instruct} on math
and code with five reward designs, then evaluate the same saved answers
with different scorers. Pass@$32$ is the share of problems where a scorer
accepts at least one of $32$ answers. On math, all five scorers find gains
of $28$ to $33$ percentage points. For any given checkpoint, their gain
estimates differ by at most $1.1$ points. On code, both judges score the
base model above $95\%$. Across three judge-trained runs, the estimated
gains range from $0.24$ to $1.43$ points. For two runs that differ only in
training seed, the weak judge gives gains of $0.48$ and $1.43$ points.
The strong judge gives both $0.71$ points. In a separate comparison of
two pretrained models, missing function names in MBPP prompts explain most
of the gap between judging answers and running their tests. On HumanEval,
which provides the function names, the strong judge scores the models
about $5$ to $9$ points higher than execution. We recommend reporting the
base score for each scorer. When a base score is close to $100\%$, small
gains call for uncertainty estimates and an independent check.
\mbox{The project page is at
{\definecolor{projectlink}{HTML}{BC5059}%
\hypersetup{urlcolor=projectlink}\urlstyle{same}%
\url{https://curiousorbitnotes.github.io/pta-cross-scorer-study/}}.}
\end{abstract}

\section{Introduction}
\label{sec:intro}

To measure whether post-training helps, we compare a model after training
with the model it started from. We call that starting model the \emph{base
model}. The result depends on how we grade the answers. A \emph{scorer} can
compare a math answer with a reference answer, run tests on generated code,
or ask a language model to judge whether a response is correct. Two scorers
can assign different verdicts to the same answer, so they can report
different gains for the same trained model.

Reinforcement learning with verifiable rewards
(RLVR;~\citealp{lambert2024tulu3,guo2025deepseekr1}) uses such checks to
give a model feedback during training. We call the score used to update the
model its \emph{training reward}. Group Relative Policy Optimization
(GRPO;~\citealp{shao2024grpo}) samples several answers to each training
prompt and uses their rewards relative to one another to update the model.
The training reward guides that update. The evaluation scorer measures
what changed after training.

We measure performance with Pass@$32$~\citep{chen2021humaneval}. Each
scorer sees the same $32$ answers for a problem. The problem counts as
solved if the scorer accepts at least one answer, and Pass@$32$ is the
fraction of problems counted as solved. For each scorer, the measured
gain is the trained model's score minus the base model's score. If a
scorer gives the base model $96\%$, its maximum possible gain is four
percentage points. We call that remaining range \emph{headroom}.
Figure~\ref{fig:math-scorers} shows how much headroom each scorer has.

Prior work asks whether reinforcement learning increases the range of
problems that a model can solve when given many attempts.
\citet{yue2025rlvr} find that improvements with few attempts can diminish
as the number of attempts grows. \citet{liu2025prorl} find gains at larger
budgets after prolonged training. These comparisons assess gains under
the evaluation rules selected for each benchmark. That leaves a separate
question open. If we give the same answers to different scorers at the same
sampling budget, do they agree on how much a model improved and how the
trained models rank?

We train Qwen-2.5-7B-Instruct~\citep{qwen2024qwen25} with five reward designs
across math and code. For each saved model, or \emph{checkpoint}, we
generate one set of answers and give it to every scorer in the comparison.
On math, all five scorers
report gains of $28$ to $33$ percentage points. They differ by at most $1.1$
points in the gain they assign to any trained model. The trained-code
comparison uses two language-model judges (Section~\ref{sec:method-eval}).
Both give the base model over $95\%$, report small gains, and rank the
models differently. The available evidence leaves the effect of training
on code performance unresolved. We also explain why an earlier test based
on which problems changed cannot distinguish training effects from
sampling noise (Section~\ref{sec:results-negctrl}).

\section{Related Work}
\label{sec:related-work}

\citet{yue2025rlvr} compare base and RL-trained models across sampling
budgets on math, code, and visual reasoning tasks. They find that training
can improve the chance of solving a problem on the first try, while the
base model can solve more problems when given many attempts.
\citet{liu2025prorl} report that prolonged training can improve performance
even at large sampling budgets, including on some problems their base
model never solves within the tested budget. \citet{li2025divergencechoice}
examine a training choice that can contribute to this pattern. In their
experiments, the usual reverse-KL penalty can narrow the range of answers
a policy produces, so Pass@$1$ can improve while performance at larger
$k$ declines. Our experiments focus on a complementary question. At a
fixed sampling budget, how much do the estimated gains and rankings
change when different scorers grade the same answers?

The training reward is also an imperfect measure of the intended task.
\citet{skalse2022rewardhacking} formalise reward hacking as cases where
improving a proxy reward reduces the true reward, a problem connected
to Goodhart's law~\citep{goodhart1981,manheim2018goodhart}.
\citet{pan2022rewardmisspec} study how this can happen when an RL agent
learns under a misspecified reward. \citet{gao2023overoptimization} find
that optimising a learned reward model too far can lower scores under a
separate reference reward. \citet{wang2026rewardhacking} review related
failure modes across language-model training methods. For GRPO with
multiple reward components, \citet{ichihara2025mogrpo} show that training
can improve one objective at the expense of others and propose
normalising the rewards to balance their contributions.
\citet{shao2025spuriousrewards} find that even random rewards can improve
math scores for some Qwen models, with weaker transfer to the other
families they test. These studies show why the chosen training signal
and the way gains are measured both deserve attention. We assign one
reward per training run and then vary the evaluation scorer.

Several studies identify sources of error in the evaluation pipeline.
\citet{hochlehnert2025soberlook} re-evaluate reasoning models under
standardised settings and show that decoding parameters, prompt format,
and random seeds can change benchmark scores. VerifyBench
\citep{yan2025verifybench} tests whether a verifier can use a reference
answer to judge a candidate response, including cases where the
candidate looks plausible but is wrong. TinyV~\citep{xu2025tinyv} addresses
the complementary error: verifiers that reject correct math answers
can deprive the model of useful feedback during training.
\citet{helff2026gamingverifiers} compare verification rules for inductive
reasoning and show that one rule rewards shortcuts that a stricter
rule rejects. In code, \citet{deshpande2026trace} find that judges detect
reward anomalies more reliably when comparing trajectories than when
reviewing them in isolation. Countdown-Code isolates when reward
hacking emerges during training~\citep{khalifa2026countdowncode}, while
EvilGenie compares several ways of detecting it in programming tasks
\citep{gabor2025evilgenie}. Other work studies training-time interventions
to reduce such behaviour~\citep{wu2026rebounds,ackermann2026gradreg}.
Our evaluation asks how much a gain changes across scorers after
training, using the same generated answers for each comparison.

Language-model judges bring a further source of disagreement.
\citet{panickssery2024selfpref} find that some judges favour their own
generations on preference tasks. \citet{wataoka2024selfprefbias} report
evidence linking self-preference to how familiar a response is to the
judge. JudgeBench~\citep{tan2025judgebench} pairs correct and incorrect
responses to test whether judges can recognise objective correctness on
challenging tasks. \citet{su2025crossingbridge} find that strong judges
given a reference answer can agree closely with human graders. We use
two judges from the same model family at different sizes and ask how
their scores for the base and trained models translate into estimated
gains. This comparison complements the call by
\citet{wu2025hiddencosts} to match testing budgets and report the design
choices that affect estimates of RLVR improvement.

\section{Method}
\label{sec:method}

Each training run uses one of the five reward functions below. After
training, we compare scorers on the same saved answers from each model.

\subsection{Five scorer designs}
\label{sec:method-rewards}

Every scorer takes a (prompt, response, gold) tuple and returns a scalar in
$[0, 1]$ that GRPO consumes as the per-rollout return.
\emph{Exact Match} (EM) returns $1$ iff a normalised extracted answer is
character-equal to the normalised gold on math. In this project's code path,
the EM-labelled implementation instead returns unit-test pass rate.
\emph{Numerical Tolerance} (NT) accepts math answers within $10^{-4}$ after
SymPy normalisation, and on code delegates every problem to EM, because its
float-tolerant code path was never implemented.
\emph{Code Execution} (CE) checks the extracted math answer for equality with
SymPy, or runs the generated program against the gold tests in a
timeout-bounded subprocess on code.
\emph{Strong LLM-as-Judge} (SLJ) uses \texttt{Qwen2.5-72B-Instruct}
served with vLLM~\citep{kwon2023vllm} at greedy decoding, emitting a
$\{0, 1\}$ verdict.
\emph{Weak LLM-as-Judge} (WLJ) uses the same stack and prompt with
\texttt{Qwen2.5-7B-Instruct}, deliberately the policy's own family, to
surface self-preference~\citep{panickssery2024selfpref}.

On code, EM, NT, and CE route through the same unit-test
implementation and form one \emph{execution family}. Their code-domain
results are withheld below.

\subsection{GRPO training}
\label{sec:method-grpo}

We use outcome-based GRPO~\citep{shao2024grpo} as implemented in TRL's
\texttt{GRPOTrainer}~\citep{vonwerra2022trl}. For a prompt $q$ with
reference answer or tests $y$, the rollout policy $\pi_{\mathrm{old}}$
samples $G=8$ completions $o_1,\ldots,o_G$. The reward function $R$ assigned
to the run scores each completed response. GRPO uses the group's mean
reward as a baseline and scales the centered reward by the group's spread:
\begin{equation}
  r_i=R(q,o_i,y), \qquad
  \bar r=\frac{1}{G}\sum_{j=1}^{G}r_j, \qquad
  A_i=\frac{r_i-\bar r}{\operatorname{std}(r_1,\ldots,r_G)+\delta},
  \label{eq:grpo-advantage}
\end{equation}
where $\delta>0$ prevents division by zero. Every token of $o_i$ shares
the same advantage $A_i$. Completions above the group mean have positive
advantages, and a group with identical rewards contributes no advantage
signal.

For token $o_{i,t}$, let $h_{it}=(q,o_{i,<t})$ be its context.
The ratios below compare its probability under the policy $\pi_\theta$
with those under the rollout policy and the frozen starting model
$\pi_{\mathrm{ref}}$:
\begin{equation}
  \rho_{it}=\frac{\pi_\theta(o_{i,t}\mid h_{it})}
                  {\pi_{\mathrm{old}}(o_{i,t}\mid h_{it})}, \qquad
  u_{it}=\frac{\pi_{\mathrm{ref}}(o_{i,t}\mid h_{it})}
               {\pi_\theta(o_{i,t}\mid h_{it})}.
  \label{eq:grpo-ratios}
\end{equation}
The PPO-clipped GRPO loss combines the advantage term and a sampled KL
penalty. Its contribution per completion token, before reduction, is
\begin{equation}
  \ell_{it}(\theta)
  =-\min\!\left\{
    \rho_{it}A_i,\,
    \operatorname{clip}(\rho_{it},1-\epsilon,1+\epsilon)A_i
  \right\}
  +\beta\left(u_{it}-\log u_{it}-1\right).
  \label{eq:grpo-loss}
\end{equation}
Here $\epsilon$ is the clipping threshold and $\beta=0.04$ sets the
strength of the penalty against the frozen model. The training reward
enters this update through $r_i$ in Equation~\ref{eq:grpo-advantage}.
We score the resulting checkpoints separately on held-out prompts.

\subsection{Experimental setup and pre-registration}
\label{sec:method-setup}

All runs start from \texttt{Qwen2.5-7B-Instruct}~\citep{qwen2024qwen25}.
Math training uses GSM8K-train~\citep{cobbe2021gsm8k} and a subset of
MATH~\citep{hendrycks2021math}. The math evaluation uses MATH-500
\citep{lightman2024letsverify} and GSM8K-test ($1819$ prompts).
Code training uses MBPP-train~\citep{austin2021mbpp} and APPS-introductory
\citep{hendrycks2021apps}. The code evaluation uses
HumanEval~\citep{chen2021humaneval} and MBPP-test ($421$ prompts, including
$164$ HumanEval and $257$ MBPP problems). The training and evaluation
splits are disjoint.

The runs use a learning rate of $1\times10^{-6}$, an effective batch of
$32$ prompts, a rollout temperature of $1.0$, AdamW with weight decay
$0.01$, and a $100$-step linear warmup. We use the checkpoint at step
$4{,}000$. The training design covers five rewards in each of the two
domains at seed $1$, with three additional runs for the seed comparisons
(\textsc{em}$\times$\textsc{math} at seeds $2$ and $3$, and
\textsc{slj}$\times$\textsc{code} at seed $2$).

Before evaluation, we committed to computing the Pass@$32$ matrix for
each checkpoint and domain before calculating rank correlations. We set
Kendall's $\tau < 0.5$ as the threshold for disagreement. For the
Reward-Type Hackability Index (RHI), we fixed the weights at
$w_1=w_2=0.4$ and $w_3=0.2$ and set a Spearman-correlation success threshold
in advance. We committed to reporting missed thresholds as such.
Section~\ref{sec:results-open} describes the two pre-registered tests that
the retained evidence cannot support.

Two setup changes departed from the pre-registration. We reduced the
code training prompt cap from $1024$ to $768$ tokens after running out of
memory on APPS prompts at $G=8$. Evaluation prompts were unaffected.
We sourced the MATH training subset from a mirror and verified its
problem IDs.

\subsection{Scoring, and why the code execution column is withheld}
\label{sec:method-eval}

For each (checkpoint, domain) pair we sample $32$ rollouts per eval prompt
once, then score them with all five scorers, giving a cross-scorer matrix in
a single round of generation. Pass@$k$ uses the unbiased estimator of
\citet{chen2021humaneval} at $k \in \{1, 8, 32\}$, with $k = 32$ as the
headline.

An audit of our own code-execution scorer, run after the first analysis of
this work was written, found four defects. HumanEval's \texttt{check()}
function was defined but never called. APPS tests were stored as a dead
literal and never executed. Six analysis scripts binarised partial test
credit at $>0.5$ and counted it as a full pass. Candidate code could return
an object with a custom \texttt{\_\_eq\_\_} that claims equality
unconditionally. Each defect can only manufacture a false pass, never a false
failure, so every affected number was an over-estimate. All four are fixed.

We do not report the corrected values, for a reason worth stating plainly. A
later revision of the same verifier added a quarantine: problems whose test
assertions cannot be safely extracted are now scored $0$ without running the
candidate at all, and our own audit notes require those problems to be
excluded from reported accuracy rather than counted as failures. Our cached
re-scoring predates that revision. Correcting for it would need a
version-stamped rescore and, because per-prompt judge verdicts were never
cached, a judge column recomputed on the same eligible subset, which we
cannot produce from the current artefacts. Publishing a second corrected
number that is one revision behind is how the first error happened. Every
code-domain \emph{accuracy} number for a trained checkpoint therefore comes
from the two LLM-judge scorers, which never routed through the execution path.

The earlier prompt-overlap claim was also computed with the execution scorer.
We therefore report none of its exact changed-prompt counts or probabilities.
Section~\ref{sec:results-negctrl} retains only the scorer-independent
mathematical reason the proposed null and direction factor are invalid.

Two exclusions follow from the same audit. Checkpoints whose \emph{training}
reward routed through the broken harness (\textsc{em}$\times$\textsc{code},
\textsc{nt}$\times$\textsc{code}, \textsc{ce}$\times$\textsc{code}) were
optimised against a signal that rewarded non-solutions, so they are not valid
instances of GRPO under those rewards and are dropped. Retraining them is
future work. The judge-trained code checkpoints
(\textsc{slj}$\times$\textsc{code} seeds $1$ and $2$,
\textsc{wlj}$\times$\textsc{code} seed $1$) were trained by an LLM judge and
were never exposed, so the code analyses use four rows: those three and the
base model. Math is untouched throughout: EM on math is string comparison,
and NT and CE on math use SymPy. The audit separately records a SymPy timeout
that can undercount correct answers.

The base-model execution numbers in Appendix~\ref{sec:appx-k128} do not rely
on the quarantined path. Base-model code samples were rescored with a hardened
harness that runs the full HumanEval prompt together with \texttt{check()} and
credits a sample only when an explicit success token follows the tests. Its
outputs are archived with checksums. A replay of two saved checkpoints
(\textsc{ce}$\times$\textsc{code} and \textsc{slj}$\times$\textsc{code} seed
$1$) produced malformed output (63.0\% and 66.9\% non-parsing), unlike the
cached samples they were first scored on. The cause is unresolved, and a companion workshop study
documents the replay in detail~\citep{fetanat2026evorobust}. No replayed
generation enters our trained-versus-base comparisons. Every trained-checkpoint
score in this paper is computed on the cached generations of the original runs.

This matters for adaptation in a plain way. A scorer that certifies an adapted
model, or that rewards an agent during training, must itself be checked
against the task: ours credited code whose tests never ran, which inflated
measured accuracy. A scorer audit with negative controls is therefore part of
the adaptation evidence, not a footnote to it. Our tasks are single-turn, so
this is a measurement prerequisite for agent evaluation, not a study of agents.

\section{Results}
\label{sec:results}

\subsection{Math: a clean negative control}
\label{sec:results-math}

\begin{table}[t]
\centering
\caption{Cross-scorer Kendall's $\tau$ point estimates on the eight math
checkpoints at Pass@$32$. All $10$ point estimates exceed the pre-registered
$\tau = 0.5$ threshold. The minimum is $0.764$. The interval qualification,
from a bootstrap over the eight checkpoints, is given in the text.}
\label{tab:tau-math}
\small
\begin{tabular}{lccccc}
\toprule
 & EM & NT & CE & SLJ & WLJ \\
\midrule
EM  & -- & $0.982$ & $0.982$ & $0.815$ & $0.982$ \\
NT  &     & -- & $1.000$ & $0.764$ & $0.929$ \\
CE  &     &     & -- & $0.764$ & $0.929$ \\
SLJ &     &     &     & -- & $0.837$ \\
WLJ &     &     &     &     & -- \\
\bottomrule
\end{tabular}
\end{table}

All ten point estimates of cross-scorer association on the eight math
checkpoints are positive and exceed the pre-registered $0.5$ threshold. The
minimum is $\tau = 0.764$ for both NT-versus-SLJ and CE-versus-SLJ
(Table~\ref{tab:tau-math}). The accompanying bootstrap ($5{,}000$ resamples)
resamples the eight checkpoints, not the $1819$ prompts. The lower endpoints
of its intervals range from $0.292$ to $1.000$, and the intervals for EM--SLJ, NT--SLJ, CE--SLJ, and SLJ--WLJ include $0.5$.
We therefore interpret the table as descriptive positive rank agreement, not
as evidence that every underlying association exceeds the threshold with high
confidence. These intervals treat the cached checkpoint scores as fixed and
do not quantify prompt-level uncertainty.

\begin{figure}[t]
  \centering
  \includegraphics[width=\linewidth]{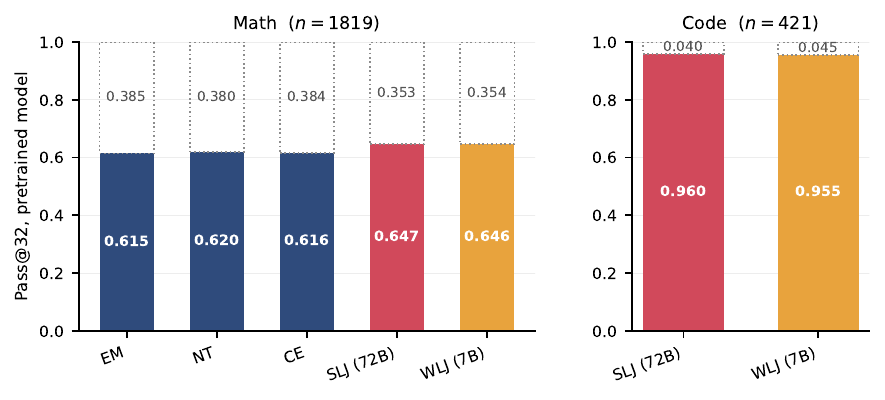}
  \caption{The pretrained model, read by every scorer we have, before
  our GRPO training. Solid bars are each scorer's Pass@$32$ for
  \texttt{Qwen-2.5-7B-Instruct}. The dotted box above each bar is the range
  that scorer has left, and carries its size. On math (left) the five
  instruments agree to within $0.032$ of each other and each retains $0.35$
  to $0.39$ of range, which is ample for the $0.28$ to $0.33$ gain they later
  attribute to GRPO. On code (right) both judges start above $0.95$ and
  retain $0.040$ and $0.045$. The gains they later report are $0.0024$ to
  $0.0143$ (positive gains only), which is $5.9\%$ to $31.6\%$ of the range
  that judge had left, each gain taken against its own judge's baseline. The
  code execution column of the original matrix is withheld
  (Section~\ref{sec:method-eval}), so code shows two scorers rather than five.
  Every bar and box here can be measured before training begins, which is
  what makes it a usable precaution rather than a post-hoc finding.}
  \label{fig:math-scorers}
\end{figure}

Agreement on ranks is the weaker statement, so we also check the effect
sizes. Every one of the seven trained math checkpoints improves on the base
under every one of the five scorers, by between $0.284$ and $0.332$
Pass@$32$, and the spread across the five scorers on any single checkpoint's
measured gain is $0.0088$ to $0.0110$
(Appendix~\ref{sec:appx-math-delta}). A practitioner who reported one of
these numbers instead of five would have been off by about $1$ point on a
$30$-point effect. The scorers also start close together on the untrained
base: EM $0.6146$, NT $0.6196$, CE $0.6163$, SLJ $0.6465$, WLJ $0.6460$, a
span of $0.032$ across five instruments of three different kinds
(Figure~\ref{fig:math-scorers}). Within these observed math checkpoints,
scorer choice changes the measured effect little. We do not infer that one
scorer is generally sufficient for math evaluation.

\subsection{Code: the judges saturate, then disagree inside the remainder}
\label{sec:results-code}

\begin{table}[t]
\centering
\caption{Code domain, Pass@$32$ under the two LLM-judge scorers, on the four
rows whose training signal never routed through the broken execution harness.
Parentheses give the difference from the base row, computed at full precision.
The last column is the disagreement between the two judges about the same
checkpoint.}
\label{tab:code-judges}
\small
\begin{tabular}{lccc}
\toprule
training reward & SLJ (72B) & WLJ (7B) & $|$SLJ$-$WLJ$|$ \\
\midrule
\textsc{base} (no GRPO) & $0.9596$ & $0.9549$ & $0.0048$ \\
\textsc{slj}$\times$\textsc{code} (s1) & $0.9667\;(+0.0071)$ & $0.9596\;(+0.0048)$ & $0.0071$ \\
\textsc{slj}$\times$\textsc{code} (s2) & $0.9667\;(+0.0071)$ & $0.9691\;(+0.0143)$ & $0.0024$ \\
\textsc{wlj}$\times$\textsc{code} (s1) & $0.9620\;(+0.0024)$ & $0.9620\;(+0.0071)$ & $0.0000$ \\
\bottomrule
\end{tabular}
\end{table}

\begin{figure}[t]
  \centering
  \includegraphics[width=\linewidth]{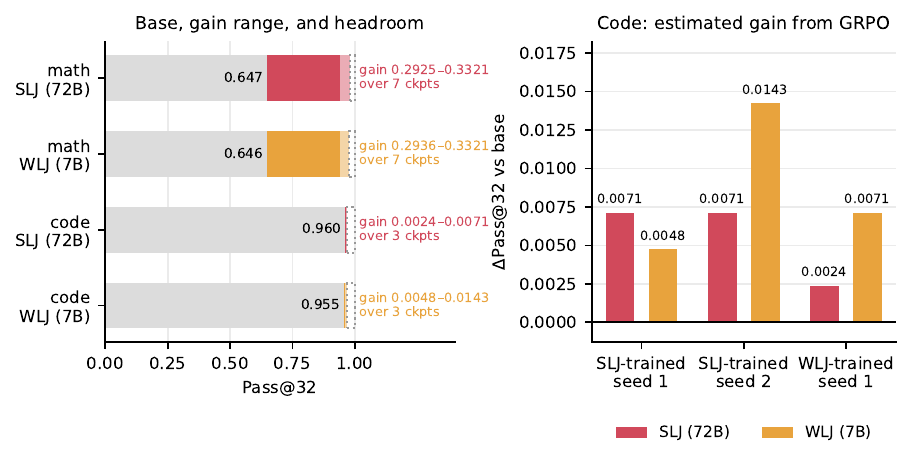}
  \caption{The left panel shows each judge's baseline Pass@$32$ in grey
  and the gain over that baseline in colour. The solid segment reaches the
  smallest estimated gain, and the lighter extension reaches the largest
  across seven math checkpoints or three code checkpoints. The right panel
  shows the gain for each retained code checkpoint. Colour identifies the
  judge that scored it, and the labels identify the training reward and
  seed. The strong judge ties the two SLJ-trained seeds, while the weak
  judge gives them different gains. Paired uncertainty intervals are
  unavailable.}
  \label{fig:judge-resolution}
\end{figure}

Both judges rate the untrained base model above $0.95$ on code
(Table~\ref{tab:code-judges}), leaving at most about $0.045$ headroom
to measure improvement. On math, the same judges start at $0.6465$ and
$0.6460$, with about $0.354$ of headroom and gains reaching $0.33$.

The judge point estimates on code are gains of $0.0024$ to $0.0143$. Since
per-prompt judge verdicts were not cached, we cannot estimate paired
uncertainty or establish that these small differences are resolved.

First, the gain assigned to different training seeds depends on the
evaluator. Only the random seed differs between the two
\textsc{slj}$\times$\textsc{code} runs, but they are separately trained
policies with different outputs. The strong judge assigns both a point
estimate of $+0.0071$. Its weaker counterpart assigns $+0.0048$ and
$+0.0143$, a spread of $0.0095$. This does not identify judge inconsistency.
It shows that a single-run point estimate depends on which trained seed and
which judge are reported.

Second, the two judges give different point-estimate orderings. The strong judge
puts the two SLJ seeds level at $0.9667$, above
\textsc{wlj}$\times$\textsc{code} at $0.9620$, above base. The weak judge
puts \textsc{slj}$\times$\textsc{code} seed $2$ first, then
\textsc{wlj}$\times$\textsc{code}, then \textsc{slj}$\times$\textsc{code}
seed $1$, then base. Those two orderings disagree about whether an SLJ-trained
checkpoint beats a WLJ-trained one, and the judges were given the same
scoring prompt, the same decoding configuration, and the same generations. Only the judge model
differs.
Without intervals, these orderings are descriptive and need not represent a
resolved difference between checkpoints.

Third, the point-estimate ordering changes with $k$. At Pass@$1$ the strong judge
ranks \textsc{slj} s1 first, at Pass@$8$ \textsc{slj} s2, and at
Pass@$32$ the two are tied (Appendix~\ref{sec:appx-passk}). The differences
being reordered are at most $0.002$.

All six trained-versus-base point estimates are positive, so a single-run,
single-judge report would look like a code improvement. The available data do
not distinguish a post-training effect from sampling uncertainty or establish
that the judge-dependent orderings are stable. We therefore report the values
descriptively and do not certify code improvement.

\subsection{Why the prompt-overlap test is invalid}
\label{sec:results-negctrl}

Our own earlier analysis of this study used overlap between changed-prompt sets from
two training seeds to argue that a small code effect was not sampling noise.
We withdraw that argument. It is invalid before any empirical control is
considered.

First, the null was wrong. At $n = k = 32$ the per-prompt
Pass@$32$ is binary, one if any of the $32$ samples passes and zero
otherwise. Under a same-model resampling null, a prompt's chance of changing
depends on its success probability and is small when it almost always or
almost never passes. Placing changes uniformly over all
prompts therefore describes the wrong process. Second, the direction factor
was vacuous: if two runs both differ from the same
binary base value, they must land on the same other value, so
direction agreement is forced rather than independent evidence.

These are mathematical properties of the statistic and do not depend on the
execution scorer's accuracy or version. Consistent with the reliability
audit, we report no exact execution-derived prompt counts or probabilities.

\subsection{Two pre-registered tests we cannot currently evaluate}
\label{sec:results-open}

We pre-registered a $\tau < 0.5$ cross-scorer disagreement test on code.
After dropping the three checkpoints trained against the broken harness, the
code domain has four rows and two usable scorers. The surviving
judge-versus-judge $\tau$ over four checkpoints ($0.548$, on scores spanning
$0.007$ and $0.014$) is descriptive only: one scorer pair over four rows, with
no prompt-level uncertainty, cannot stand in for the pre-registered
comparison. We report the test as not evaluable rather
than substituting one that happens to pass.

We also pre-registered a Reward-Type Hackability Index, a weighted z-score of
entropy drift, KL to base, and output-length drift over the first $500$
steps, with success defined by the pre-registered Spearman-correlation
threshold against end-of-training proxy-gold divergence across the ten
primary runs. That target variable is
recorded as exactly zero for the three code runs trained under the broken
harness, and for no other code run. A flat zero at exactly the three
compromised runs is a signature of the defect rather than a measurement, and
the pre-registered correlation is computed across all ten runs. Both
the original verdict and any post-hoc rescue of it are unavailable until
those runs are retrained. Our failure protocol commits us to reporting a
missed threshold rather than retrofitting one. We extend it here to a target
that cannot be measured.

\section{Discussion}
\label{sec:discussion}

The baseline score tells us how much improvement a scorer can register.
On math, the five scorers have room to detect gains and report similar
improvements. The code judges start near $0.96$, leaving less than $0.05$
on the scoring scale. The weak judge's two-seed spread of $0.0095$
Pass@$32$ is about a fifth of that range. The strong judge also scores a
second pretrained model at $0.969$ on code
(Appendix~\ref{sec:appx-k128}). Baseline scores therefore show when there
is little headroom. Detecting gains within that range also requires
paired uncertainty and evaluation with an independent scorer.

Disagreement can reveal a mismatch between what a task asks for and what
its tests require. In the separate MBPP run, adding the expected function
name to the prompt raised execution Pass@$128$ from $0.125$ to $0.794$ for
Qwen and from $0.167$ to $0.817$ for Llama
(Appendix~\ref{sec:appx-k128}). HumanEval gives the function name in the
prompt and has a smaller judge--execution gap. Looking at the benchmarks
separately shows how much the pooled comparison depends on the task
specification. When a judge and an execution scorer disagree, a useful
first check is whether the prompt states the interface the tests expect.

An aggregate gain also hides which problems changed. Two trained models
can show the same net improvement by gaining and losing on different
problems. Saving the base and trained verdicts for each problem makes
those changes visible under each scorer. A paired analysis can then
show whether scorers notice gains on the same problems or whether the
reported gains depend on a small set of disagreements.

Work on reasoning coverage varies the sampling budget
\citep{yue2025rlvr,liu2025prorl}. Our main comparison varies the scorer
at $k=32$. The math gains are consistent across the five scorers at
that budget. On code, judge-trained checkpoints show slightly higher
lexical diversity than the base on distinct-$1$ ($0.0730$ to $0.0755$
versus $0.0718$), distinct-$2$, and the fraction of unique completions.
These statistics describe the text of the sampled completions. Claims
about broader reasoning coverage require evidence about which problems
the completions solve.

The verifier defects in Section~\ref{sec:method-eval} invalidated an
earlier code result. When training and evaluation use the same verifier,
both can reward the same incorrect response. An independent scorer can
help detect such failures. Reporting the baseline, results by training
seed, and paired uncertainty makes the strength of the evidence easier
to assess.

\section{Conclusion}
\label{sec:conclusion}

For math, all five scorers report similar GRPO gains on the same saved
completions. The estimated Pass@$32$ gain is about $0.30$ and varies by
at most $0.011$ across scorers for any checkpoint. On code, the two
judges give different rankings near the top of the scoring scale.
Without paired uncertainty estimates or execution scores for the trained
checkpoints, the effect of GRPO on code remains unresolved.
The same comparison can be applied to existing training runs with saved
answers. Scoring those answers again at the same sampling budget shows
whether the reported gain holds under another evaluation rule.
We recommend reporting the base-model score for each scorer before
interpreting a post-training gain. When the baseline is close to the
ceiling, a stronger test would rerun the excluded code training
configurations with a validated verifier, retain per-prompt verdicts
for both base and trained checkpoints, and test whether the gain holds
under independent evaluation.

\section{Limitations}
\label{sec:limitations}

The retained code comparison covers one base model and three judge-trained
checkpoints. Execution scores for the trained checkpoints were withheld
after the verifier check and an unresolved mismatch between replayed and
original generations (Section~\ref{sec:method-eval}). The original
per-prompt judge verdicts are unavailable, which prevents paired intervals
for the code gains. The math rank-correlation intervals resample the eight
checkpoints and leave prompt-level uncertainty unmeasured.
We studied GRPO post-training on one base model in two domains. The
independent Pass@$128$ analysis covers pretrained models only
(Appendix~\ref{sec:appx-k128}).

\clearpage
\bibliographystyle{plainnat}
\bibliography{references}

\appendix

\section{Code judge scorers at Pass@1 and Pass@8}
\label{sec:appx-passk}

Table~\ref{tab:appx-passk} gives the four valid code rows under both judges
at $k \in \{1, 8\}$. The saturation is milder at small $k$ and the ordering
is unstable across $k$: the strong judge ranks \textsc{slj} s1 above
\textsc{slj} s2 at $k = 1$, below it at $k = 8$, and level at $k = 32$. Every
difference being reordered is at most $0.002$.

\begin{table}[h]
\centering
\caption{Code domain under the two judge scorers at Pass@$1$ and Pass@$8$,
valid rows only. Parentheses give the difference from the base row.}
\label{tab:appx-passk}
\small
\begin{tabular}{lcccc}
\toprule
 & \multicolumn{2}{c}{Pass@$1$} & \multicolumn{2}{c}{Pass@$8$} \\
\cmidrule(lr){2-3}\cmidrule(lr){4-5}
run & SLJ & WLJ & SLJ & WLJ \\
\midrule
\textsc{base}                          & $0.8396$ & $0.8141$ & $0.9385$ & $0.9320$ \\
\textsc{slj}$\times$\textsc{code} (s1) & $0.8548\;(+0.0152)$ & $0.8242\;(+0.0100)$ & $0.9466\;(+0.0081)$ & $0.9387\;(+0.0067)$ \\
\textsc{slj}$\times$\textsc{code} (s2) & $0.8529\;(+0.0133)$ & $0.8220\;(+0.0079)$ & $0.9468\;(+0.0083)$ & $0.9379\;(+0.0060)$ \\
\textsc{wlj}$\times$\textsc{code} (s1) & $0.8539\;(+0.0143)$ & $0.8247\;(+0.0106)$ & $0.9454\;(+0.0069)$ & $0.9383\;(+0.0063)$ \\
\bottomrule
\end{tabular}
\end{table}

\section{Math trained-versus-base deltas under all five scorers}
\label{sec:appx-math-delta}

\begin{table}[h]
\centering
\caption{Change in Pass@$32$ relative to the untrained base for each math
checkpoint under each scorer, with the across-scorer spread. Every cell is a
gain, and no checkpoint's measured gain moves by more than $0.011$ depending
on who measures it.}
\label{tab:appx-math-delta}
\small
\begin{tabular}{lcccccc}
\toprule
run & EM & NT & CE & SLJ & WLJ & spread \\
\midrule
\texttt{ce\_math\_s1} & $0.2969$ & $0.2963$ & $0.2908$ & $0.3018$ & $0.3002$ & $0.0110$ \\
\texttt{em\_math\_s1} & $0.3304$ & $0.3282$ & $0.3227$ & $0.3310$ & $0.3321$ & $0.0093$ \\
\texttt{em\_math\_s2} & $0.2925$ & $0.2930$ & $0.2875$ & $0.2941$ & $0.2963$ & $0.0088$ \\
\texttt{em\_math\_s3} & $0.2903$ & $0.2897$ & $0.2842$ & $0.2925$ & $0.2936$ & $0.0093$ \\
\texttt{nt\_math\_s1} & $0.2908$ & $0.2908$ & $0.2859$ & $0.2969$ & $0.2941$ & $0.0110$ \\
\texttt{slj\_math\_s1} & $0.3293$ & $0.3277$ & $0.3222$ & $0.3310$ & $0.3293$ & $0.0088$ \\
\texttt{wlj\_math\_s1} & $0.3293$ & $0.3271$ & $0.3216$ & $0.3321$ & $0.3304$ & $0.0104$ \\
\bottomrule
\end{tabular}
\end{table}

\section{A second pretrained model, and a larger sampling budget}
\label{sec:appx-k128}

The body reports one pretrained model at $k \leq 32$. This appendix adds a
second model family and $k = 128$, for the pretrained models only. Trained
checkpoints are excluded here for the reason given in
Section~\ref{sec:limitations}.

Execution scores here come from a hardened harness applied to all $421$
problems. It runs the full HumanEval prompt followed by \texttt{check()}, runs
the MBPP assertions after a fixed import preamble, and credits a sample only
when an explicit success token is reported after the tests. It passes every
reference solution and fails every negative control we tried. The judge
columns are the strong judge's per-sample verdicts on the same generations.
The generations come from a separate run with $128$ samples per prompt
(temperature $1.0$, top-$p$ $0.95$, at most $1{,}024$ new tokens), so Pass@$32$
here is estimated from $n = 128$ rather than read from the $n = 32$ matrix of
the body. The submitted version of this appendix used an earlier verifier on
$406$ of the problems, with the judge columns on all $421$. The values below
replace those.

\begin{table}[h]
\centering
\caption{Pretrained models under hardened execution and the strong judge, on
all $421$ code problems. The last two columns split the gap by benchmark: the
HumanEval prompts name the function under test, while our MBPP prompts do not.}
\label{tab:appx-k128}
\small
\begin{tabular}{llccccc}
\toprule
pretrained model & $k$ & execution & SLJ (72B) & gap & HumanEval gap & MBPP gap \\
\midrule
\texttt{Qwen-2.5-7B-Instruct}  & $32$  & $0.4288$ & $0.9610$ & $0.5321$ & $0.0677$ & $0.8285$ \\
\texttt{Qwen-2.5-7B-Instruct}  & $128$ & $0.4418$ & $0.9715$ & $0.5297$ & $0.0549$ & $0.8327$ \\
\texttt{Llama-3.1-8B-Instruct} & $32$  & $0.4305$ & $0.9688$ & $0.5383$ & $0.0921$ & $0.8230$ \\
\texttt{Llama-3.1-8B-Instruct} & $128$ & $0.4679$ & $0.9857$ & $0.5178$ & $0.0610$ & $0.8093$ \\
\bottomrule
\end{tabular}
\end{table}

The pooled gap is $0.52$ to $0.54$ for both families at both budgets, but it
should not be read as judge over-credit. Nearly all of it comes from MBPP,
whose prompts in our evaluation set do not state the name of the function
that the tests call. Strict execution therefore fails answers that solve the
task under another name: about $94\%$ (Qwen) and $87\%$ (Llama) of failing
MBPP samples end in a \texttt{NameError} on the expected name. Renaming the
candidate function after the fact lowers the pooled Pass@$128$ gap to $0.119$
(Qwen) and $0.112$ (Llama). A separate generation run with the name added to
the MBPP prompt raises base MBPP Pass@$128$ from $0.125$ to $0.794$ (Qwen) and
from $0.167$ to $0.817$ (Llama). Both adjustments are sensitivity checks, not
replacements for the benchmark. On HumanEval, where the prompt fixes the
signature, the strong judge exceeds execution by $0.055$ to $0.092$. At
Pass@$128$ the $95\%$ bootstrap intervals of that gap, resampling the $164$
HumanEval prompts, are $[0.018, 0.098]$ (Qwen) and $[0.024, 0.098]$ (Llama). The
submitted version read the pooled gap as the judge over-crediting both
families by the same margin, and we withdraw that reading. What carries over
from the body is the saturation: the strong judge rates both pretrained
families above $0.95$ at Pass@$32$ on code, so neither leaves much room for a
judge-measured gain.

On math the comparable gap for \texttt{Llama-3.1-8B-Instruct}, strong judge
minus exact match, is $0.038$ at both $k = 32$ and $k = 128$. Only exact match
and the two judges were run on Llama math, so this is a three-scorer check
rather than a five-scorer one. Llama's math scores are themselves high (strong
judge $0.952$ at Pass@$32$), so the math-versus-code headroom contrast of the
body does not carry over to Llama unchanged.

\section{Reproducibility}
\label{sec:appx-repro}

Math results and the code judge scores in the main text come from the
cross-scorer Pass@$k$ matrices. Appendix~\ref{sec:appx-k128} uses the hardened
execution harness, whose per-sample outputs are archived with checksums. The
separate $n = 128$ base-model run retains the judge verdicts needed for its
intervals. The original matrix lacks per-prompt judge verdicts, so its code
gains have no paired intervals. The audit log dates the verifier corrections,
and the original execution scores are archived separately. The claim ledger
links reported values to their analyses and records withdrawn claims. The
checkpoint replay discrepancy remains unresolved (Section~\ref{sec:limitations}).

\end{document}